\documentclass[5p,twocolumn]{elsarticle}

\usepackage{amsmath}
\usepackage{graphicx}
\usepackage{amssymb}
\usepackage{color,soul}
\usepackage[T1]{fontenc}
\usepackage{bm}
\usepackage{ae,aecompl}
\usepackage{nameref}

\biboptions{numbers,sort&compress}

\begin{document}

%\preprint{}

%\title{`Memristor' models that are not memristive: }

\title{On the validity of memristor modeling in the neural network literature}

\author[1]{Yuriy V. Pershin\corref{cor1}}
\ead{pershin@physics.sc.edu}

\author[2]{Massimiliano Di Ventra}
\ead{diventra@physics.ucsd.edu}

\cortext[cor1]{Corresponding author}
\address[1]{Department of Physics and Astronomy, University of South Carolina, Columbia, South Carolina 29208, USA}
\address[2]{Department of Physics, University of California, San Diego, La Jolla, CA 92093, USA}


\date{\today}

\begin{abstract}
An analysis of the literature shows that there are two types of non-memristive models that have been widely used in the modeling of so-called ``memristive'' neural networks.  Here, we demonstrate that such models have nothing in common with the concept of memristive elements: they describe either {\it non-linear resistors} or certain {\it bi-state systems}, which all are devices {\it without} memory. Therefore, the results presented in a significant number of publications are at least questionable, if not completely irrelevant to the actual field of memristive neural networks.
\end{abstract}

%\keywords{}

\maketitle

This Letter refers to a number of  publications on ``memristive'' neural networks (MNNs) published during the last decade~\cite{T1_1,T1_2,T1_3,T1_4,T1_5,T1_6,T1_8,T1_9,T1_10,T1_11,T1_12,T1_13,T1_14,T1_15,T1_16,T1_17,T1_18,T1_19,T1_20,T1_21,T1_22,T2_1,T2_2,T2_3,T2_4,T2_5,T2_6,T1_7,T2_7,T2_8,T2_9,T2_10,T2_11,T2_12,T2_13,T2_14,T2_15,T2_16,T2_17,T2_18,T2_19,T2_20,T2_21,T2_22,T2_23,T2_24,T2_25,T2_26,T2_27,T2_28,T2_29,T2_30,T2_31,T2_32,T2_33,M1_1,M1_2,M1_3,M1_4,M1_5}
 (note that this list may be incomplete, as there may be other publications that slipped through our search). We put the word ``memristive'' in quotes, because as we will show in the present paper, the referenced
 published papers refer to models that have nothing to do with resistive memories (memristive elements).

 In fact, in Refs.~\cite{T1_1,T1_2,T1_3,T1_4,T1_5,T1_6,T1_8,T1_9,T1_10,T1_11,T1_12,T1_13,T1_14,T1_15,T1_16,T1_17,T1_18,T1_19,T1_20,T1_21,T1_22,T2_1,T2_2,T2_3,T2_4,T2_5,T2_6,T1_7,T2_7,T2_8,T2_9,T2_10,T2_11,T2_12,T2_13,T2_14,T2_15,T2_16,T2_17,T2_18,T2_19,T2_20,T2_21,T2_22,T2_23,T2_24,T2_25,T2_26,T2_27,T2_28,T2_29,T2_30,T2_31,T2_32,T2_33,M1_1,M1_2,M1_3,M1_4,M1_5},
 two types of {\it non-memristive} models were used in the modeling/simulation of MNNs. Our main statement in this work is that the devices considered in these publications have {\it no memory} of past dynamics, and as such they cannot  represent memristive elements. Consequently, the results obtained with these models have no relevance to the field of {\it actual} memristive neural networks~\cite{Pershin2010}.

To simplify the presentation, we will refer to the aforementioned models as ``type 1'' and ``type 2'' models. The type 1 model~\cite{T1_1,T1_2,T1_3,T1_4,T1_5,T1_6,T1_8,T1_9,T1_10,T1_11,T1_12,T1_13,T1_14,T1_15,T1_16,T1_17,T1_18,T1_19,T1_20,T1_21,T1_22} claims to
approximate a ``memristive element'' by an expression of the type
\begin{equation}\label{eq:1}
  R_M^{(1)}(\dot{V}_M(t))=\left\{
  \begin{array}{ccc}
    R_{on}, & \dot{V}_M(t)>0 & \\
    R_{off}, &  \dot{V}_M(t)<0 & , \\
    \textnormal{unchanged}, &  \dot{V}_M(t)=0 &
  \end{array}
  \right.
\end{equation}
where $R_M^{(1)}$ is supposed to be the memristance (memory resistance), $V_M(t)$ is the voltage across the device, $R_{on}$ and $R_{off}$ are the low- and high-resistance states of the device, respectively, and the dot denotes the time derivative. To the best of our knowledge, the first use of Eq.~(\ref{eq:1}) was proposed in Ref.~\cite{T1_13}.

In the type 2 model~\cite{T2_1,T2_2,T2_3,T2_4,T2_5,T2_6,T1_7,T2_7,T2_8,T2_9,T2_10,T2_11,T2_12,T2_13,T2_14,T2_15,T2_16,T2_17,T2_18,T2_19,T2_20,T2_21,T2_22,T2_23,T2_24,T2_25,T2_26,T2_27,T2_28,T2_29,T2_30,T2_31,T2_32,T2_33},
the memristance in a MNN is represented by an expression of the form
\begin{equation}\label{eq:2}
  R_{M,ij}^{(2)}(V_j)=\left\{
  \begin{array}{ccc}
    \hat{R}_{ij}, & |V_j|>T_i,\\
    \check{R}_{ij}, &  |V_j|<T_i, \\
  \end{array}
  \right.
\end{equation}
where $T_i$ are thresholds, $\hat{R}_{ij}$ and $\check{R}_{ij}$ are constants, and $V_j$ is the voltage at a node $j$ of the network.

Based on a literature search, the model represented by Eq. (\ref{eq:2}) was pioneered by the authors of Ref.~\cite{T2_15}.
Moreover, there is a sub-set of publications~\cite{M1_1,M1_2,M1_3,M1_4,M1_5} where both type 1 and type 2 models are mentioned. While Eqs. (\ref{eq:1}) and  (\ref{eq:2}) look different, they have a  feature in common: the devices that they describe are {\it not} memristive elements.
\begin{figure*}[t]
	(a)\centering{\includegraphics[width=75mm]{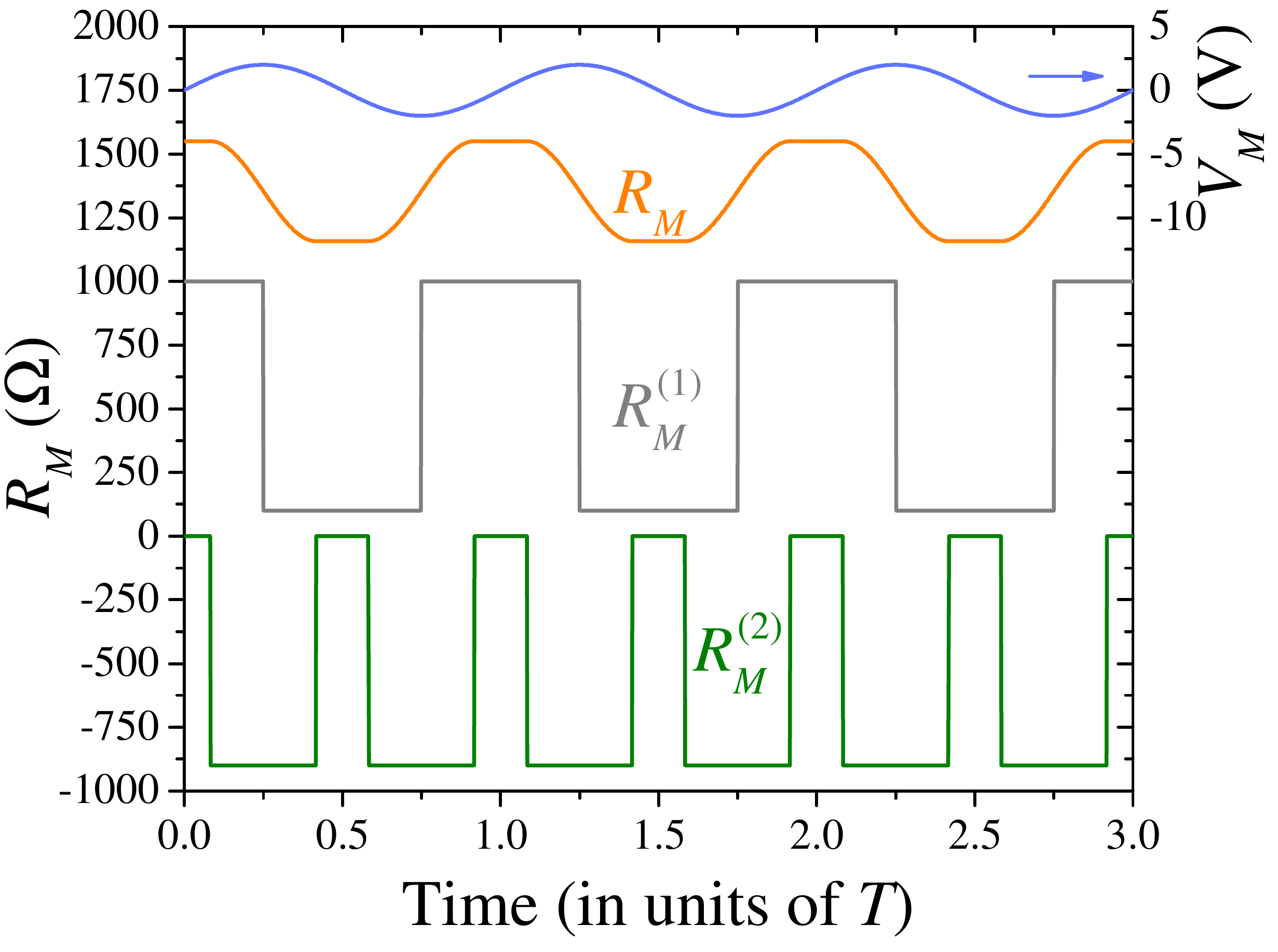}}
	(b)\centering{\includegraphics[width=75mm]{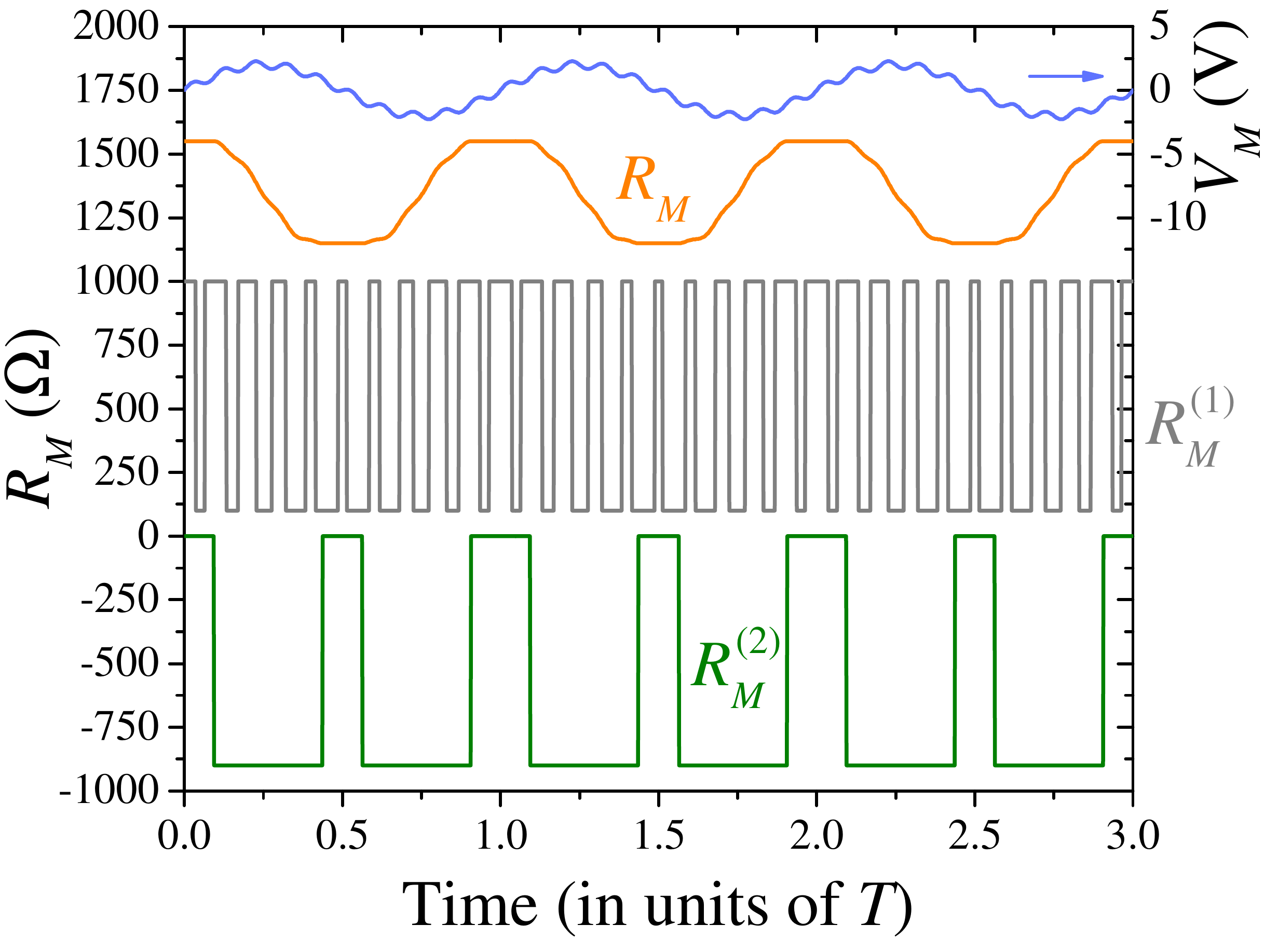}}
	\caption{Memristance as a function of time calculated using the type 1 [Eqs. (\ref{eq:1})], type 2 [Eqs. (\ref{eq:2})], and an actual memristive model [Eqs. (\ref{eq:3}) and (\ref{eq:4})]. The curves have been shifted for clarity by 1~k$\Omega$. The top curves in (a) and (b) represent the applied voltage given by $V_M(t)=2 \sin (2\pi t/T)$~V and $V_M(t)=2\sin (2\pi t/T)+0.3 \sin (2\pi t/(0.1T))$~V, respectively. These plots were obtained using the following set of parameter values: (type 1 device) $R_{on}=100$~$\Omega$ and $R_{off}=1$~k$\Omega$; (type 2 device) $\hat{R}_{ij}=100$~$\Omega$ and $\check{R}_{ij}=1$~k$\Omega$; (threshold-type memristive system~\cite{pershin18a})  $R_{on}=100$~$\Omega$, $R_{off}=1$~k$\Omega$, $V_t=1$~V, and $\beta T=1800$~k$\Omega$V$^{-1}$; $R_M=R_{off}+(R_{on}-R_{off})x$; $\dot{x}=\textnormal{sign}(V_M)\beta(|V_M|-V_t)$ if $|V_M|>V_t$, and $\dot{x}=0$ otherwise.}\label{fig:1}
\end{figure*}

To proceed, let us first recall the definition of {\it actual} memristive elements~\cite{chua76a}. These are two-terminal resistive devices with memory defined (in the voltage-controlled case~\cite{chua76a}) by
\begin{eqnarray}
I&=&R^{-1}_M\left( \bm{x}, V_M \right) V_M, \label{eq:3} \\
\dot{\bm{x}}&=&\bm{f}\left(\bm{x}, V_M \right), \label{eq:4}
\end{eqnarray}
where $I$ and $V_M$ are the current through and voltage across the device, respectively, $R_M\left( \bm{x}, V_M\right)$ is the memristance (memory resistance), $\bm{x}$ is an $n$-component vector of internal state variables, and $\bm{f}\left(\bm{x}, V_M\right)$ is a vector function.

The memory feature of memristive elements is related to their internal state that evolves according to Eq. (\ref{eq:4}) and is manifested in the device response (notice that $R_M$ is a function of $\bm{x}$). When subjected to time-dependent input, memristive elements typically exhibit pinched hysteresis loops. Importantly, due to the presence of memory, these loops must be strongly dependent on the input frequency (and voltage amplitude)~\cite{chua76a,diventra09a}. Note that this is physically necessary for {\it any} system with memory~\cite{13_properties}.
For instance, for high-frequency input signals the hysteresis loop closes, as  there is not enough
time for the internal state variables to follow the fast-varying input.

%and only survives at appropriate frequencies that are characteristic of the time decay of the memory.

Now, a brief comparison of Eqs. (\ref{eq:1}) and (\ref{eq:2}) with Eqs. (\ref{eq:3}) and (\ref{eq:4}) is sufficient to establish the fact that the devices described by the type 1 and type 2 models are {\it not} memristive. While the actual memristive elements are characterized by a memory (time non-locality) of signals applied in the past,  the response of  type 1 and type 2 devices is effectively {\it history-independent}. This feature is readily evident in the case of type 2 model that simply describes a {\it non-linear resistor}, whose resistance is fully determined by the {\it instantaneous} voltage (which, in some publications~\cite{T2_15}, is not even the voltage across the device).

In the case of type 1 models, the instantaneous response is determined by the sign of the time-derivative of the voltage. Even though the time derivative implies the dependence on the voltage at an infinitesimally close preceding moment of time, this alone is not sufficient for the device to be classified as a memristive
element. We emphasize that not only does the time derivative of the voltage not enter Eqs. (\ref{eq:3}) and (\ref{eq:4}), but also it is difficult to imagine an
actual {\it physical} device with such a voltage differentiation capability (definitely the physical memristive elements behave differently~\cite{pershin11a}).

Finally, consider the last line in Eq. (\ref{eq:1}), which is the condition that the response of type 1 devices is unchanged when  $\dot{V}_M(t)=0$. Such an isolated point condition is irrelevant since it is singular.

\begin{figure}[t]
	\centering{\includegraphics[width=75mm]{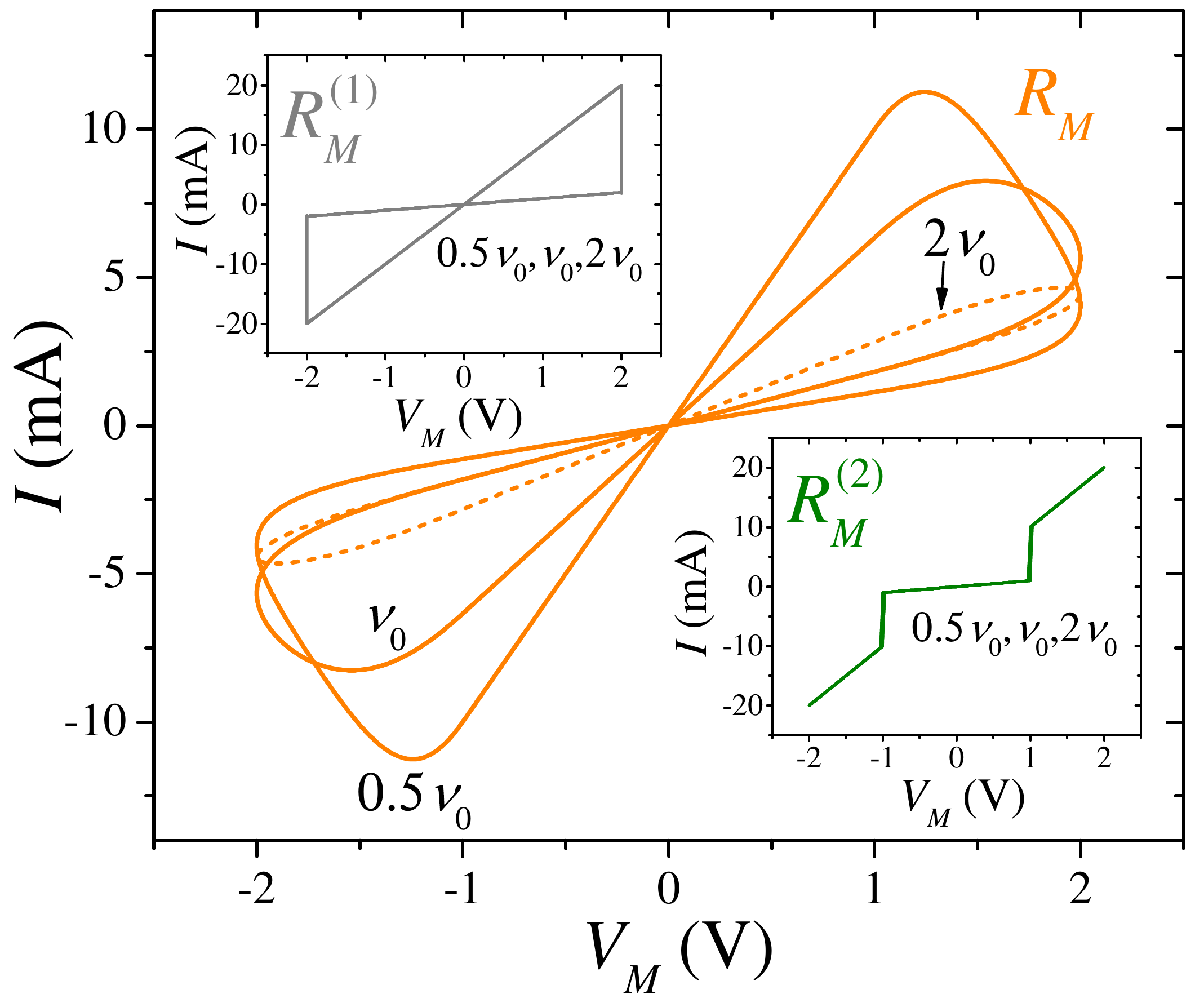}}
	\caption{Current-voltage characteristics of device models considered in this Letter. Frequency-dependent pinched hysteresis loops of an actual memristive model (in the center) contrast with the {\it frequency-independent} loops of type 1 model (top left inset) and {\it non-hysteretic} characteristics of type 2 model (bottom right inset). This plot was obtained using the same model parameters as in Fig. \ref{fig:1}; $V_M(t)=2 \sin (2\pi \nu_i t)$~V; $\nu_0=1/T$.} \label{fig:2}
\end{figure}

To further emphasize the distinction between the type 1 and type 2 devices with an actual memristive model, Fig.~\ref{fig:1} compares their response under the condition of periodic bias. Here, the memristive device is exemplified by a threshold-type model~\cite{pershin18a,pershin09b} that mimics the most common bipolar memristive elements~\cite{pershin11a}, while the response of the type 1 and 2 devices is plotted based on Eqs.~(\ref{eq:1}) and (\ref{eq:2}),
respectively.

First of all, consider the application of a simple sinusoidal voltage. This is shown in Fig.~\ref{fig:1}(a). The response of the type 1 device seems deceptively
similar to that of an actual memristive element, but close inspection shows that such a similarity is superficial.
Indeed, unlike the actual memristive element, the type 1 device exhibits {\it frequency-independent} pinched hysteresis loops in the voltage-current plane (shown in the top left inset in Fig. \ref{fig:2}) and its switching occurs {\it always} at voltage extrema but not at the threshold voltages defined by the physical processes responsible for memory as in actual memristive elements. Frequency-independence of the I-V curve is also evident for the type 2 device
as shown in the bottom right inset in Fig. \ref{fig:2}. In addition, the non-hysteretic character of these curves indicates the absence of memory in
 the type~2 model.

%~\footnote{The frequency dependence of hysteresis loops is one of the basic properties of memristors~\cite{chua76a}.},

Next, consider the response to more complex waveforms. Fig.~\ref{fig:1}(b) shows that small higher-frequency oscillations added to the main sinusoidal waveform change drastically the response of the type 1 device. Now its
resistance switches at the frequency of small-amplitude signal, and has nothing in common with the behavior of an actual memristive element
(whose resistance has not changed significantly compared to Fig.~\ref{fig:1}(a)). This demonstrates that the type 1 devices are highly sensitive to small
amplitude variations as opposed to the actual memristive element. In Fig. \ref{fig:1}(a) and (b), the resistance dynamics for the type 2 model involves a frequency doubling. According to the discussion above, the absence of memory in this model is evident.

To conclude, in this Letter we have shown that two types of ``memristive'' models widely used in the literature to model/simulate memristive neural networks are, in fact, {\it not} memristive. During the past decade, multiple studies based on these models have been reported in leading specialized journals, such as Neurocomputing, Neural Networks, etc. There are serious reasons to doubt the validity of these papers as the models adopted by their authors do not qualify as memristive, and as such have nothing to do with actual memristive neural networks.
\bibliographystyle{elsarticle-num}
\bibliography{memr_mm}

\end{document}